# Retrieval-Augmented Large Language Models as Components of Cognitive Computing architecture for Regulatory Knowledge Management

Dariusz Nowak-Nova[1]

[1]WSB University, Poland
ORCID: 0000-0002-7556-5677
e-mail: dnowaknova@gmail.com

ABSTRACT

**The aim of this article is to verify whether integrating large language models (LLMs) with the Retrieval-Augmented Generation (RAG) architecture enables their transformation from standalone generative models into components of cognitive computing infrastructure with enhanced epistemic reliability. The study proposes an architectural approach based on locally deployed LLMs operating in on-premises environments without high-end GPU accelerators and examines their applicability in supporting regulatory management processes requiring continuous analysis and interpretation of legal acts. The proposed solution combines local LLMs with external knowledge repositories, creating a hybrid cognitive architecture in which the language model performs semantic interpretation while the RAG layer provides controlled knowledge retrieval, contextualization, and traceability of information sources. The implementation was validated using the Ollama and LM Studio execution environments together with the Polish language models Bielik and PLLuM running on consumer-class hardware. The results demonstrate that augmenting LLMs with RAG significantly improves the factual consistency, domain specificity and normative precision of generated texts while reducing the risk of unsupported content generation. Furthermore, the study shows that integrating RAG introduces auditability, controlled knowledge management and dynamic updating of regulatory information without retraining the language model. The findings indicate that locally deployed LLMs enhanced with RAG should be regarded not merely as text generation tools but as semantic processing modules within cognitive computing infrastructures supporting regulatory compliance and organizational decision-making in environments characterized by high legal and informational volatility.**

## 1. INTRODUCTION

Contemporary Large Language Models (LLMs) are increasingly viewed not only as advanced algorithmic systems for processing textual data and identifying complex patterns, but also as highly specialized modules that form the building blocks of Cognitive Computing (CC) architectures. This becomes viable through the application of architectures using Retrieval-Augmented Generation (RAG), in which the inference model, in order to find fragments that best answer the user's prompt, works in conjunction with the layer responsible for searching and dynamic matching of search results from external databases of knowledge, documents or API. This allows for overcoming the limitations of ready-made LLMs in the production-ready version, increasing the readability, accuracy and contextuality of the outputs (Brown et al., 2025; Khan et al., 2024). Although the adoption of such solutions in the business sector is at an experimental level, the possibilities of employing RAG in the field of legal practice (Barron et al., 2025; Ho et al., 2025; Karp et al., 2026), in the management of governance and audit processes (Xu et al., 2025), the advantages of launching it in search for corporate procedures or internal regulations (Chen et al., 2025) are indicated. As noted by Arslan, Munawar and Cruz, despite the significant development of research on the integration of RAG with LLM, there is a discrepancy between applications of innovative nature and the practical use of these technologies. The authors argue that the integration of RAG into LLM, in particular based on Natural Language Processing (NLP), offers sufficient economic rationale for research focusing on the development of applications using RAG in the areas of extraction and acquisition of information necessary for making informed decisions (Arslan et al., 2024). Generating an extended search, improving the effectiveness of lexical search methods, is based on the corpora of structured documents retrieved from outside the model environment. These documents often constitute a business secret, and exploiting publicly available LLMs requires the use of services available in an external cloud-based infrastructure. This may potentially pose a risk to privacy or a risk of information leaks (Cheng et al., 2025; Huly et al., 2024). The computing environment outside the organizational domain, in the calculation model based on the infrastructure of a third party, is burdened with significant limitations. Since the contents of legal documents, internal regulations or organizational information must be sent outside the organization's infrastructure, there occurs a problem of maintaining data sovereignty, which hampers full control over its processing and further use. This entails a direct risk of regulatory non-compliance (e.g. in terms of the protection of legally protected information, business secrets or sensitive data), as well as a limited ability to demonstrate where and under what conditions the data is processed. The lack of full control over the model and its changes may also be a significant disadvantage – the cloud provider may update models or its parameters without user control, which in an environment requiring interpretive stability may lead to difficulties in maintaining the comparability of analyses over time. The operational dependence on an external provider also remains an important limitation, including the availability of the service, pricing policy, prompt limits or changes in license terms. Finally, in applications requiring LLM integration with internal knowledge repositories and RAG, the cloud introduces additional communication delays (Huang et al., 2025).

The use of LLMs in the organization's on-premises computing infrastructure is free from these disadvantages and is an adequate solution for systems that require full control over the configuration and evolution of language models. It also provides opportunities for deep integration with local resources and with internal information systems, where epistemic control, process auditability and regulatory compliance are crucial. The basic advantage of such an approach is the preservation of information sovereignty, understood as the possibility for an organization to exercise control over the location of data processing, their life cycle and the way content is used in analytical and decision-making processes. The processing of normative documents, internal materials and organizational data within its own infrastructure domain minimizes the risk of confidentiality violations, reduces exposure to undesirable cross-border transfers and facilitates the demonstration of compliance with legal regimes regarding the protection of information and sensitive data (Alvarez et al., 2025; Bumgardner et al., 2024; Cook et al., 2023).

In the source literature, it is quite common to compare advanced local LLM platforms. It is noted that with the groundbreaking GPT-OSS releases from OpenAI, DeepSeek-V3 from Shēndù Qiúsuǒ, Qwen3-Coder-480B from Alibaba Cloud, Llama 4 from Meta and Gemma 3 from Google, local LLM systems have begun to match the performance of cloud-based services while ensuring full data privacy. They offer advanced inference, multimodal support, agent-oriented coding capabilities, and built-in tool calling. This makes local LLMs both efficient and safe (Joshi, 2025; Phogat et al., 2025; Puspitasari et al., 2025; Song et al., 2025).

This also applies to the most recognizable Polish initiatives, which include Bielik and PLLuM models, developed as open or partially open solutions intended for research and institutional applications. Polish LLMs are a specialized class of models trained or tuned on corpora covering the Polish language, including administrative, legal, journalistic and scientific texts. Their main feature is a much better reproduction of inflection, syntax and semantic phenomena characteristic of the Polish language compared to general models, trained primarily on English-language data (Ociepa et al., 2025;

Sułkowski and Ulatowska, 2025; Ulatowska, 2025).

The article attempts to verify whether combining large language models with the RAG architecture transforms LLM from closed production-ready tools with established parameters into cognitive computing components with increased epistemic reliability. This should limit the phenomenon of generating content ungrounded in sources ("hallucinations"). From the perspective of analysis and management systems, this will mean the possibility of significantly increasing factual precision and reducing the risk of hallucinations, which pose one the most serious threats to the application of language models in decision-making environments.

## 2. RESEARCH METHOD

### 2.1. GENERATIVE TEXT MODELS EMPLOYED

Two Polish models were tested: Bielik in the `Bielik-1.5B-v3.0-Instruct-GGUF` version and PLLuM in the `PLLuM-12B-nc-250715` version. Both models are available in a standard format compatible with the `Hugging Face` ecosystem, which means that their launch is based on the toolchain intended for `Transformer`-type models.

In industrial practice, the Linux system environment is required in order to launch them, in a research and development configuration, a Linux/macOS/Windows environment with `Python` installed is sufficient. The study used the macOS environment with the Apple M4 processor and 24 GB of unified Metal 4 memory compliant with the current Metal Performance Shaders (MPs) framework.

In order to use the models as an element of the RAG architecture, it was necessary to run additional components in parallel. This included a model generating representation of non-numeric data (`BERT` adapted to Polish was employed), a local vector database storing representations of document fragments (`ChromaDB` was used) and a prompt orchestration layer for coordinating search processes and output generation (`LangChain` was used). Taking into account, the use of models as a component of cognitive infrastructure, it became necessary to additionally implement a model serving layer in the form of a local inference server, which allows parallel handling of multiple prompts (`vLLM` was used). The libraries used to parse OCR documents in the repositories were `Pypdf/PyMuPDF`, `sentence-transformers`, `faiss-cpu`, `python-docx`, `LangChain`, `numPy`.

### 2.2. RUNTIME ENVIRONMENT FOR RAG

Ollama and LM Studio were used as the runtime environment for the application of RAG. Both Ollama and LM Studio are tools for a local LLM launch supporting various operational scenarios. Ollama acts as a light service layer operating in the LLM service mode, providing a programming interface for eliciting inference. The application architecture is aimed at integration with other applications and data processing pipelines. The work model is based on automation, scripts and a stable API, thanks to which it can perform the role of a local inference server in on-premises architectures. LM Studio was designed as an environment for interactive work with various language models. It is focused on the operation of models, prototyping user `prompts` and assessing the quality of the outputs. LM Studio supports RAG mechanisms and work with local document catalogues in a way integrated with the user interface, without the need to build a separate programming infrastructure. In technical terms, it allows for indicating local file resources as sources of knowledge. These resources can then be automatically processed in order to prepare internal index structures. During the interaction with the user, LM Studio implements a query processing scheme typical for the RAG architecture, which involves an initial search for relevant fragments in the uploaded documents, and then feeding the selected context as part of the input to the language model. This allows language models to be quickly linked to local knowledge repositories.

Access to online legal resources was an important issue. Runtime environments do not have a direct capability to communicate outside of locally launched LLMs. They are an interface for communication with the model responsible for generating text and possibly running other programmes. This interface can be extended using a uniform, standard protocol that extends the functionality of LLM models. This uniform extension mechanism (protocol) is the Model Context Protocol (MCP). LM Studio version 0.3.39 has been integrated with the Zotero reference manager version 8.0.2 as an external context source. In architectural terms, LM Studio played the role of an MCP client and an inference engine, while the MCP Zotero server provided structured access to the publication database (metadata, abstracts, attached PDF files). Technically, MCP Zotero played the role of a specialized retrieval layer, and LM Studio implemented the generative and interpretative part. The architecture was not a classic vector-based RAG, but a tool-based retrieval mechanism in which the selection of documents was carried out not through embeddings search, but through the Zotero tool interface allowing direct use of existing, curated libraries, along with their collection structure, descriptions and expert notes. The local repository reaching the size of 22.84 GB contained 17 377 indexed documents.

In the case of Internet addresses, LM Studio did not download the content of websites independently, but was embedded in the architecture, in which the download and extraction of content from a URL was carried out by the built-in Tavily MCP server running a specialized Internet search engine designed for large language models in the LM Studio graphic interface. This platform was treated as an auxiliary retrieval component, in which LM Studio played only the role of the generation engine. The Tavily MCP server, as a source of context, implemented a `web-fetch` service consisting in the download of content from a website (HTML), clearing it of presentation elements, and then returning the structured text and automatically injecting it into the input context of the model, in which LM Studio used the returned fragments as a context for output. The sources of legal information used were the *Internetowy System Aktów Prawnych/Internet System of Legal Acts* (https://isap.sejm.gov.pl), *Dzienniki Ustaw/Journals of Laws* (https://dziennikustaw.gov.pl/DU), *Monitor Polski/Official Gazette of the Republic of Poland,* (https://monitorpolski.gov.pl/MP), *Official Journals* (https://dziennikiurzedowe.gov.pl), *Rządowy proces legislacyjny/Government legislative process* (https://legislacja.gov.pl) and the *Korpus Dyskursu Parlamentarnego/Parliamentary Discourse Corpus* (https://kdp.ipipan.waw.pl/query_corpus/5/).

### 2.3. The Course of the Study

The study was conducted in two stages. In the first one, locally hosted LLMs were interviewed using the `llama.cpp`. The Ollama application, which is a "friendly interface" for `llama.cpp`, was abandoned due to the better performance of executed instructions. Then the queries were repeated using LM Studio, in which the LLM model answers were combined with semantically selected RAG outputs from the external corpus. This was performed in three sequences:

1) **Retrieval phase:** The system converted the user `prompt` to an `embeddings`. In this process, each semantically separated fragment is converted into a dense floating-point vector using a local embedding model supported by the model. Then, the system compares the query with an indexed collection of vectorized documents consisting of semantically segmented texts relevant to a given domain. The vectors consisted of metadata at the fragment level (e.g. file name, page number, position) embedded in thematic collections of a local `ChromaDB` instance. In this way, on the basis of similarity metrics, `k` of the most similar segments were indicated.
2) **Extension phase:** The `k` segments retrieved from the vector store in order to construct an extended prompt, were merged into the user's `prompt`.
3) **Generation phase:** The model has generated a homogeneous and extended output based on both information retrieved from the model and `k` segment of the response to the `prompt`.

Four research questions were asked:

**RQ1:** Provide the definition of "real estate" in the Polish legal system.
**RQ2:** Which laws contain this definition?
**RQ3:** Which legal acts have developed this definition?
**RQ4:** Nowhere else?

Two rounds of research were carried out. In the first round, 4 `prompts` were fed to the engines of both LLMs. In round two, the LM Studio interface, which utilised an expanded RAG query, was included in the query entry phase. Then, the texts generated were subsequently evaluated using automated assessment tools designed to determine whether the texts were grammatically correct, readable, comprehensible, and what their characteristics were. The evaluation employed Jasnopis.pl, a readability assessment application, and Logios.dev, a syntactic analysis tool. Both applications are based on computational linguistics and are intended to assess text comprehensibility, readability, and stylistic properties. Jasnopis.pl measures text comprehensibility using a seven-point scale, where 1 denotes a very easy text and 7 denotes a very difficult/scientific text. Logios.dev functions as an editor-type application, focusing on the analysis of language in terms of syntax and difficulty (fuzziness). Both tools measure the difficulty of the text (fuzziness), analyse the length of sentences and words. They do not carry out any substantive assessment.

## 3. RESULTS

The study results are presented in the tables below. For enhanced readability and ease of comparison, the generated texts have been positioned side by side.

**Table 1.** *Texts generated using production-ready LLMs*

| Bielik | PLLuM |
|---|---|
| **RQ1: Provide the definition of "real estate" in the Polish legal system** | |
| Real estate is a plot of land, a building, or part of a building permanently attached to a plot of land.<br>Under the Polish legal system, real estate is classified into several categories: land property (plots of land), building property, and premises constituting separate property (buildings or parts thereof that constitute separate objects of ownership). real estate is defined as a material object occupying a specific area of space. The owner of real estate has the right to use, dispose of, and derive benefits from their own property, provided that such use does not infringe the rights of others or violate provisions of law. Real estate is registered in land and mortgage registers, which keep information about the legal status of the property and its encumbrances and limitations. | Real estate is everything that is permanently attached to a given plot of land or is a component of this plot. This is a term used in civil law and refers to any material objects that are not movable.<br>Real estate includes, among other things, buildings, residential and commercial premises, as well as the common areas of multi-unit residential buildings.<br>Real estate may be owned by natural or legal persons and be subject to various legal regulations applying to its acquisition, transfer or inheritance. |
| **RQ2: Which laws contain this definition?** | |
| The definition of real estate in the Polish legal system is found primarily in the Civil Code, which regulates issues related to ownership and other rights in rem. Specifically, the definition of real estate is contained in Art. 46 § 1 and Art. 48 of the Civil Code:<br>Art. 46 [Profits]<br>§ 1. Real estate is a part of the land surface that constitutes a separate object of property (land), as well as buildings permanently connected with the land or parts of such buildings, if they constitute a separate object of property from the land under special provisions.<br>Art. 48 [Component part]<br>Subject to the exceptions provided for in the act, buildings and other facilities permanently attached to the land are considered component parts of the land.<br>In addition, matters relating to the registration of real estate are further regulated by the Act on Land and Mortgage Registers and on Mortgage (6 July 1982), which establishes the rules governing land and mortgage registers and their role in trading real estate.<br>It is also worth mentioning the Real Estate Management Act (of 21 August 1997), which regulates the principles of managing real estate of the State Treasury and local government units.<br>Collectively, these ,laws constitute the legal framework governing real estate in Poland. | The definition of real estate is contained in the Polish Civil Code.<br>Pursuant to Art. 46 §1 of this legal act: "Real estate is part of the earth's surface constituting a separate object of ownership (land), as well as buildings permanently attached to the land or parts of such buildings if, under specific regulations, they constitute an object of ownership separate from the land."<br>In addition, it is worth mentioning that this definition can be developed and supplemented by other legal acts and jurisprudence. |
| **RQ3: Which legal acts have developed this definition?** | |
| In the Polish legal system, in addition to the Civil | The definition of real estate contained in the Civil |

| Bielik | PLLuM |
|---|---|
| Code, the Act on Land and Mortgage Registers and on Mortgage, and the Real Estate Management Act, there are a number of other legal acts developing the definition of real estate in various contexts:<br>The Construction Law Act (of 7 July 1994) sets out the principles of design, construction and maintenance of buildings. Although it is not the primary act governing the definition of real estate as such, it influences the characteristics and the manner in which buildings utilise real estate.<br>The Spatial Planning and Development Act (27 March 2003) regulates the processes of spatial planning. It may affect the value and purpose of real estate depending on its location and the specificity of certain areas in terms of urban planning.<br>The Water Law Act (18 July 2001) defines the legal status of waters and regulations regarding the use of waterfront land or land lying in the vicinity of rivers or lakes. It may affect the characteristics and use of certain types of real estate, especially those linked to water.<br>The Municipal Economy Act (8 March 1990) concerns the activities of municipalities relating to publicly owned property. This may be relevant when defining or using certain types of local government properties.<br>These legal acts do not alter the core definition of real estate contained in the Civil Code but develop it in the context of specific situations and types of real estate. | Code constitutes the primary legal source on this issue.<br>However, additional legal acts further develop the concept and introduce specific regulations relating to real estate. For instance, the Real Estate Management Act of 21 August 1997 establishes the legal framework governing real estate transactions and procedures associated with its subdivision, consolidation and expropriation.<br>In addition, it is worth mentioning the Construction Law Act, which regulates issues related to the construction and maintenance of buildings and affects the perception of real estate as a functional whole. |
| **RQ4: Nowhere else?** | |
| Within the Polish legal system, the principal regulations governing real estate are contained primarily in the Civil Code, the Act on Land and Mortgage Registers and on Mortgage, and the Real Estate Management Act. These legal acts form the basis for the definition, registration and trading of real estate.<br>As regards specific issues relating, for example, to the management of jointly held real estate in residential buildings, they may be regulated by the Act on Ownership of Premises (24 June 1994). However, the definition of real estate as such is primarily contained in the aforementioned legal acts.<br>It should also be noted that there may exist other special laws (lex specialis) which regulate specific aspects pertaining to real estate in specific contexts. However, the basic definition and general principles are mainly contained in the aforementioned legal acts. | The definition of real estate can also be developed by jurisprudence.<br>In practice, this means that in the case of disputes regarding the legal status of a given object or its part, judgments issued by Polish courts prevail.<br>Moreover, it is worth mentioning the views held within civil law doctrine, which may influence the interpretation of the concept of real estate and its practical application in specific real-life or business situations. |

*Source:* the author's own elaboration based on research

**Table 2.** *Texts generated using LLMs extended with RAG*

| Bielik+RAG | PLLuM+RAG |
|---|---|
| **RQ1: Provide the definition of "real estate" in the Polish legal system** | |
| In the Polish legal system, real estate is defined as a part of the earth's surface, constituting a separate object of ownership. This is a fundamental element of the regulations on the right in rem and is crucial | In the Polish legal system, the concept of real estate is defined in the Civil Code (CC). Pursuant to Art. 46 §1 of the Civil Code, "Real estate is part of the earth's surface constituting a separate object of ownership |
| for the economic turnover and the financial system of the country. Real estate may take the form of land property, building property, or premises (residential or commercial). Land is defined as undeveloped areas or areas developed with structures permanently attached to the ground. Buildings may constitute a separate real estate if they are erected on the land that is the subject of perpetual usufruct or if the right to the building results from a separate arrangement. On the other hand, premises (residential and commercial) are separate parts of buildings that can be legally traded regardless of the land or the remaining part of the building. This definition is of fundamental importance to all real estate transactions and is the basis for establishing and protecting property rights. | (land), as well as buildings permanently attached to the land or parts of such buildings if, under specific regulations, they constitute an object of ownership separate from the land." This definition is of a general nature and forms the basis for more detailed regulations pertaining to real estate. It is worth noting that Polish law distinguishes various types of real estate: Land - all parts of the earth's surface that constitute a separate object of ownership (e.g. construction and agricultural plots). Buildings permanently attached to land - structures erected on the ground that cannot be easily moved. Parts of such buildings – where specific statutory provisions stipulate so (e.g., residential units marked as separate ownership). Real estate may also be movable if it is a component part of land or building property. An important element of the definition is permanent attachment to the land, meaning that it is not possible to easily move a given object to another place without destroying its utility value. |
| **RQ2: Which laws contain this definition?** | |
| The definition of "real estate" and regulations related to real estate are contained mainly in two basic legal acts: the Civil Code (CC) and the Real Estate Management Act.<br>- Civil Code – Art. 46 § 1 of the Civil Code defines real estate as a part of the earth's surface, constituting a separate object of ownership. In addition, Art. 235 of the Civil Code defines buildings and other structures permanently attached to the land as constituent parts of the real estate (unless specific regulations provide otherwise).<br>- Real Estate Management Act – This law regulates issues related to the expropriation, division and consolidation of real estate and its sale by the State Treasury or local government units. Although it does not directly define the term "real estate", it refers to various aspects of real estate management and trading, which is important in the context of civil law.<br>Additionally, detailed regulations regarding individual types of real estate (e.g. land, building, premises) may be found in other laws and implementing regulations, such as the Surveying and Cartographic Law or the Act on Ownership of Premises. | Answering the question about real estate regulations in the Polish legal system requires detailed reference to several key legislative acts. The most important ones are presented below, along with their relevant articles and clauses: Civil Code (CC) Article 46 § 1: "Real estate is part of the earth's surface constituting a separate object of ownership (land), as well as buildings permanently attached to the land or parts of such buildings if, under specific regulations, they constitute an object of ownership separate from the land." Surveying and Cartographic Law Article 2 point 8: Definition of a cadastral parcel as an area unit used to describe the legal status of real estate. Real Estate Management Act Chapter I, Art. 4: It contains general rules for real estate transfer and management in Poland, as well as the definition of real estate as an object of ownership right. Act on Land and Mortgage Registers and on Mortgage Article 2 point 1: It defines real estate as land with component parts and buildings permanently attached to that land, if they constitute a separate object of ownership. Spatial Planning and Development Act Article 2 point 13: It defines land as a real estate intended for development or used in a specific way, in accordance with the local spatial development plan. Act on the Protection of Agricultural and Forest Land Article 4 point 1: It introduces the concept of real estate as land intended for agriculture or forestry, which is subject to special legal protection. Local Taxes and Fees Act, Article 2 para. 1: It defines the subject of taxation in the form of real estate as land, buildings or parts thereof, as well as structures related to conducting business activity. Each of these laws contains provisions regulating various aspects of real estate - from definitions, through the rules of transfer and use, to tax issues and the protection of agricultural and forest land. |
| **RQ3: Which legal acts have developed this definition?** | |
| The definition of real estate contained in the Civil | The development of real estate definition in the |
| Code (CC) has been developed and elaborated by a number of other legislative acts that regulate various aspects related to real estate. The most important ones are presented below:<br>- Act on Land and Mortgage Registers and on Mortgage – It regulates the keeping of land and mortgage registers, which are the basic register of rights to real estate. This law defines the rules for entries and deletions as well as the protection of property laws on real estate.<br>- Real Estate Management Act – It defines the principles of managing public real estate (State Treasury and local government units), including expropriation, division, collation and sale.<br>- Surveying and Cartographic Law – It regulates issues related to the registration of land and buildings (real estate cadastre), which is crucial for determining the legal status of real estate.<br>- Act on Ownership of Premises – It specifies the rules for establishing and managing separate ownership of premises (residential and commercial), which is crucial for residential real estate.<br>- Spatial Planning and Development Act – It regulates issues related to the land use, which affects the possibility of utilising the property (e.g. construction, exploitation).<br>- Act on the Protection of Agricultural and Forest Land – It defines the principles of protecting agricultural and forest lands and transforming them into non-agricultural or non-forest purposes.<br>These legal acts, in conjunction with the Civil Code, form a comprehensive system of regulations pertaining to real estate in Poland. | Polish legal system results from a number of legislative acts that adapt the general principles contained in the Civil Code to the specificity of various types of real estate and regulate the detailed aspects of their use and management. The key legal acts that have expanded the definition of real estate are presented below: Real Estate Management Act Chapter I: Establishes the basics of real estate management in Poland. Article 4 point 1: It contains a general definition of a land real estate as part of the earth's surface constituting a separate object of ownership. Act on Land and Mortgage Registers and on Mortgage Article 2 point 1: It specifies that a real estate may consist of land with component parts and buildings permanently attached to that land. Spatial Planning and Development Act Article 2 point 13: It defines the concept of land as real estate intended for development or used in a specific way, in accordance with the local spatial development plan. Construction Law Act Article 3 point 2: It defines buildings and their parts as separate construction objects constituting real estate or its component part. Local Taxes and Fees Act Chapter III, Art. 1a para. 1: It indicates the diversity of taxation objects in the form of lands, buildings and parts thereof, as well as structures related to conducting business activity. Agriculture Act Article 4 point 1: It defines agricultural real estate as land intended for conducting agricultural activity, which is subject to specific regulations regarding transfer and use. Environmental Protection Act Article 3 point 48: It contains a definition of real estate as an area that may be subject to activities that affect the natural environment. Each of these laws makes an important contribution to the development of the general definition of real estate contained in the Civil Code, adapting it to the characteristics of various areas of law and practical needs related to real estate management and trading. |
| **RQ4: Nowhere else?** | |
| In the Polish legal system, the basic definitions and regulations regarding real estate are contained mainly in the Civil Code and in several key laws that expand this subject in detail. The aforementioned legal acts (the Civil Code, the Act on Land and Mortgage Registers and on Mortgage, the Real Estate Management Act, Surveying and Cartographic Law, the Act on Ownership of Premises, the Spatial Planning and Development Act and the Act on the Protection of Agricultural and Forest Land) constitute the legal basis for issues related to real estate.<br>Apart from this, there are also other industry regulations or specific implementing regulations that may address specific aspects of real estate management (e.g. property taxes, building law in the scope pertaining to buildings). However, the main legal framework for the definition and regulation of real estate is already comprehensively covered in the abovementioned acts.<br>More specific or specialized issues can be developed | In the Polish legal system, the definition of real estate is developed not only in legislative acts but is also reflected in numerous court rulings and interpretations of administrative bodies. The key sources that further extend this definition are presented below: Supreme Courts Jurisprudence, many real estate cases examined by the Supreme Court (SC) have contributed to clarifying the concept of real estate in Polish law. Examples of important decisions of the Supreme Court include: Resolution of 15 February 2019 (case number III CZP 83/18): It explains that a building permanently attached to the land may constitute separate real estate only if specific regulations provide so. Judgment of 4 February 2022 (case number II CSKP 103/22): It confirms that parts of a building cannot be considered real estate without explicit statutory regulation. Interpretations by administrative bodies the Ministry of Development and Technology and the Ministry of Finance regularly issue interpretations regarding the |
| in the legal doctrine (jurisprudence, commentaries), but the normative acts themselves no longer contain additional definitions beyond those basic ones. | definition of real estate in the context of various aspects of the law, such as taxes or spatial planning. Examples of such cases include: Individual interpretation of the Ministry of Finance of 15 January 2023 (No. 0114-KDIP1-2.4012.719.2022.2.MC): It explains that for tax purposes, real estate consists not only of the land and building, but also of the component parts attached thereto. Statement of the Ministry of Development and Technology of 8 February 2024: It emphasizes the importance of local spatial development plans in defining the use of the land as real estate for construction or other purposes. It is also worth devoting attention the doctrinal works and academic publications that analyse the concept of real estate from the perspective of civil, administrative and tax law. Thanks to a wide range of legal sources, the definition of real estate in the Polish legal system remains flexible and adapted to the changing economic and social realities. |

*Source:* the author's own elaboration based on research

Table 3 and Table 4 present the results of the linguistic analyses of the texts generated. The Gunning FOG Index (FOG) used in both applications is an indicator determining the accessibility of a text, indicating the number of years of education required to understand it. The lower the score, the more accessible the text is. This method is based on the analysis of the average length of sentences and the percentage of complex (sophisticated) words. The Logios.dev application additionally employs the PLI index (Plain Language Index), which takes into account 10 language features and shows the extent to which the text complies with the *plain language* standard as a percentage.

**Table 3.** *Study using the Jasnopis.pl application*

| Parameter studied | Bielik | PLLuM | Bielik+RAG | PLLuM+RAG |
|---|---|---|---|---|
| Text difficulty level | 6 | 5 | 7 | 5 |
| Gunning FOG Index | 13.93 | 12.25 | 15.48 | 13.16 |
| Number of words | 561 | 278 | 599 | 928 |
| Average word length [syllables] | 2.39 | 2.33 | 2.54 | 2.33 |
| Average sentence length [words] | 15.6 | 21 | 15.9 | 21 |
| Average paragraph length [words] | 26 | 84 | 63.5 | 84 |
| Percentage of sophisticated words | 3.0 | 1.8 | 5.5 | 1.8 |
| Percentage of nouns | 42.5 | 34.9 | 34.7 | 34.9 |
| Percentage of sophisticated nouns | 10.6 | 6.0 | 6.3 | 6.0 |
| Percentage of verbs | 7.5 | 11.5 | 12.6 | 11.5 |
| Percentage of sophisticated verbs | 0.4 | 0 | 0 | 0 |
| Percentage of adjectives | 22.1 | 25.9 | 30.7 | 25.9 |
| Percentage of sophisticated adjectives | 2.7 | 3.0 | 9.5 | 3.0 |

*Source:* the author's own elaboration based on research

**Table 4.** *Study using the Logios.dev application*

| Parameter studied | Bielik | PLLuM | Bielik+RAG | PLLuM+RAG |
|---|---|---|---|---|
| Gunning FOG Index | 14 lat | 13 lat | 15 lat | 13 lat |
| PLI Index | 4% | 4% | 2.6% | 0% |
| Impersonality | 5.9% | 7.3% | 5.2% | 6% |
| Finite verbs | 3.3% | 6.9% | 3.4% | 3.8% |
| Verbs | 13% | 19% | 16% | 13% |

*Source:* the author's own elaboration based on research

## 4. STUDY FINDINGS

The study examined whether integrating large language models with the RAG architecture transforms LLM from closed tools with fixed parameters in the production-ready version into components of a cognitive infrastructure with enhanced epistemic reliability.

First of all, it should be noted that both in the study performed in Jasnopis.pl and Logios.dev, the FOG index values for all variants are in the range of about 12 to 15 years of education. This means that the texts analysed – regardless of the model and mode of generation – are specialised and require at least a secondary or early academic level of education. This is consistent with the nature of the texts generated, which refer to the concepts of civil law and use legal terminology. There is a noticeable difference between the variants of the content generated from the production-ready version and models with RAG, at the same time there is no simple rule stating that the content in Table 1 is simpler than in Table 2 - the division in this respect applies to the LLM models themselves. It can be noted that variants with lower FOG (PLLuM and PLLuM+RAG – 12.25 and 13.16 and 13 years at Logios.dev, respectively) are more comprehensible, in contrast to the variants Bielik and Bielik+RAG, which reach the level of 13.93 and 15.48 and 13-15 years of education. At the same time, the PLLum+RAG variant from Table 2 is at the level of the production-ready version of Bielik from Table 1. This confirms the assumption that RAG technology increases the share of specialist terminology responsible for the length of words (average number of syllables per word reaching 2.54 in the Bielik+RAG variant) and a higher percentage of sophisticated words. For RAG variants, the FOG index also increases. In the light of the definition of the FOG indicator, a higher value means a greater difficulty of the text. In the context of the texts analysed, this is not a defect, but a desirable feature, because a text that precisely uses legal terminology and describes normative relations must inevitably show a higher level of specialization.

From the perspective of differences between the examined texts, the parameter "percentage of sophisticated words" is particularly important. It can be concluded that the variants generated by Bielik are characterized by a more normative style, the share of sophisticated words is significantly higher in them (5.5% in Bielik+ RAG vs. 1.8% in PLLuM variants). In Bielik+RAG, the average word length increases (2.54 syllables vs. 2.39) and the percentage of sophisticated words increases significantly (3.0% → 5.5%). At the same time, the share of sophisticated adjectives grows (2.7% → 9.5%). A similar dependence occurs in relation to sophisticated nouns. In the study conducted, Bielik+RAG provided texts strongly saturated with legal terms, which are characterized by a significantly higher lexical load. At the same time, structural analysis showed that the PLLuM style is associated with longer sentences and paragraphs. The variants PLLuM and PLLuM+RAG are characterized by an average sentence length of 21 words and very long paragraphs (84 words on average), which corresponds to the method of building long enumerations and sequences of descriptions for legal regulations. It can be concluded that the accumulation of editorial units and administrative as well as legal terms in PLLuM and PLLuM+RAG performs a referential rather than a definitional function. Nevertheless, it can be confirmed that the RAG versions contain greater saturation of the text with specialist terminology and more precise normative descriptors.

From the point of view of language quality, the proportions of the parts of speech are also important. The texts generated show a very high share of nouns (over 40% in Bielik and over 34% in the others), with a very low share of verbs at the same time. In combination with a high percentage of sophisticated adjectives (especially 9.5% in Bielik+RAG), this leads to the nominalisation of statements, which is a typical feature of legal texts. The impersonality index in Table 4 is higher in the variants PLLuM and PLLuM+RAG, and the share of finite verbs is at a similar level in all variants (about 3.4%), with a clearly higher value (6.9%) in the case of PLLuM. The noticeable large reduction (up to 3.8%) of this parameter in the case of PLLuM+RAG is an interesting phenomenon. It may mean that together with the RAG environment, the PLLuM model in the production-ready version moves away from a purely nominal style, in which free narrative dominates, towards sentence structures expressing normative relations.

Particularly important is the PLI indicator used in the Logios.dev app. Values at the level of 0% to 4% indicate that all the texts analysed meet the plain language criteria to a very small extent (maximizing message simplification). This confirms that the texts generated can be considered as developed under strict conditions of normative styles, with a large number of references to legal acts and a high density of specialist terms characteristic of administrative and normative language.

Based on the data from Tables 3 and 4, it can be confirmed that the versions of texts generated using the RAG mechanism are qualitatively better than the versions generated without RAG support, and this superiority does not mean content simplification, but a higher level of terminological control, thematic consistency and style appropriate for legal texts.

When analysing the substantive correctness, it was noted that the content generated by the PLLum+RAG model contains an erroneous fragment. In Table 2, in question *RQ1: Provide the definition of "real estate" in the Polish legal system*, the following statement appears in the last paragraph: "*Real estate may also be movable if it is a component of land or building property*." This wording is in conflict with the basic construction of the right in rem and the definition contained in the Civil Code (Journal of Laws of 1964 no. 16, item. 93). Movable property and real estate constitute disjoint categories of objects in property law. Component parts of real estate do not acquire the character of movable objects, but on the contrary – they become part of real estate, with the latter being the principal object. Pursuant to Art. 191 of the Civil Code, the ownership of real estate extends to a movable object which has been attached to real estate in such a manner that it has become its component part. This principle implies that whatever is erected or planted on the land shares the legal fate of this land. The consequence of such a legal status will be the extension of property rights to the movable object attached (Krzyszczak, 2020). Therefore, this applies to the right of ownership rather than changing a "movable object" into "real estate". An error of this kind should be deemed fundamental, since it touches upon the very essence of the concept of real estate.

The manner in which the PLLum+RAG model-generated content presents the role of respective special acts can also be deemed problematic. Reading the answer to the question *RQ3: Which legal acts have developed this definition?* one may get the impression that in the Polish legal order there exists laws that have "*expanded the definition of real estate*" (Table 2, question RQ3) containing autonomous definitions of real estate in the civil law sense. Meanwhile, acts such as *the Spatial Planning and Development Act* (Journal of Laws 2023 no. 80, item 717), *Local Taxes and Fees Act* (Journal of Laws 1991 no. 9, item 31) or *the Act on the Protection of Agricultural and Forest Land* (Journal of Laws 1995 no. 16 item 78) use functional concepts, adapted to the purposes of a given regulation, and do not redefine the concept of real estate as an object of property law. The lack of distinction between the civil law definition and the concepts used for the purposes of administrative, tax or planning law leads to a methodological confusion of normative orders. This is a significant substantive flaw involving the lack of a distinction between the definition of the concept and its use in special regulations.

## 5. Discussion

Law is not merely a set of textual provisions, but a multilayered system of norms, values, relationships and contextual references, which is why the analysis of legal acts is one of the most complex interpretative tasks. In contrast to simple information documents, the legal text is characterised by high semantic density, precise logical structure and strict embedding in the hierarchy of law sources. This means that it cannot be interpreted in a linear fashion, through classical methods of information extraction, but it is necessary to use an approach drawing on cognitive processes. A legal provision does not state a fact but establishes a rule of action, imposes an obligation, or prohibits a specific behaviour. Comprehending content of this kind requires interpretation that includes not only literal analysis but also systemic, teleological, or functional construction — all these methods appeal to human cognitive mechanisms, such as the capacity for contextual reconstruction, forecasting regulatory impacts, assessing consequences, and categorising exceptions and exemptions. Such tasks may be carried out by Cognitive Computing systems.

Linking LLMs with the RAG architecture is one of the most accessible solutions in the business environment enabling the transformation of LLMs from closed tools with fixed parameters in the production-ready version into cognitive infrastructure components. The main advantage is the introduction of a transparent, external layer of knowledge, which acts as a long-term memory of the system and is a direct source of content used in the output generation process. Unlike models operating solely on the knowledge encoded in the model parameters, the LLMs coupled with RAG semantic search mechanisms operates on current, controlled and domain-specialised information resources. This enhances the capability of precisely adapting the system to the specificity of the domain. Through the appropriate selection and curation of the document corpus, structures of indices and embedding models, it is possible to shape the system's semantic space in a way that reflects the terminology, concepts and relationships characteristic of a given field. As a result, the language model, despite remaining a general purpose one, begins to function in a strictly specialised cognitive context. Owing to this, it becomes possible not only to recreate the bases of interpretation, but also to employ the system in the formal procedures of audit, control and expert verification. In contrast to the classic applications of LLM, in which the sources of knowledge remain implicit, the RAG architecture allows for treating the content generated to as an outcome of a controlled cognitive process, and not as an autonomous production of a model. It is also possible to dynamically update knowledge resources without having to re-train the model. In regulatory systems, in which the content of applicable standards changes frequently, the RAG architecture allows for the almost immediate inclusion of amendments, new legal acts or interpretations by updating the repository of documents and their vector representations. As a result, the LLMs is able to operate on the current state of normative knowledge, while simultaneously maintaining the stability of its own architecture and parameters.

From the point of view of CC architectures, it is particularly important that RAG introduces a clear separation between the language processing mechanism and the mechanism of knowledge storage and selection. In such an architecture, the LLM serves as an interpretive and semantic module, while the *retrieval* layer is responsible for the system's information perception, i.e. for the identification of knowledge fragments relevant to a given task-oriented context. Such a separation fosters the construction of hybrid architectures, in which the LLM is not an independent carrier of knowledge, but instead becomes an element of a wider cognitive system that encompasses memory, control and validation.

However, a comparison of the LLM with the CC paradigm reveals a fundamental difference between these systems. CC stems from the tradition of cognitive science and symbolic artificial intelligence, where the goal is to create machines capable of replicating cognitive processes. LLMs were created in the deep learning movement as statistical models whose task is to predict the next elements of language sequences based on massive datasets. LLMs constitute an extremely efficient yet passive source of semantic information processing which enables one to operate on texts and meanings without the need to directly process low-level linguistic structures. In CC, knowledge is structural - rules, ontologies, semantic networks, and models are explicit and undergo manipulation. The system "knows" that certain concepts are subtypes of others, that certain relationships are causal, and others are normative or logical. In the LLM, knowledge is encoded sub symbolically, in the form of neural network weights. The model does not have access to concepts as such, does not distinguish between facts, rules and hypotheses, nor does it consciously operate on them. As a result, its answers may be linguistically correct and logically coherent, yet result not from an actual manipulation of meanings, but rather from the replication of statistical correlations present in the training data. The study demonstrated conceptual errors of one of the LLMs, such as confusing the categories of real estate and movable property or ambiguous attribution of definitional functions to legal acts that, in reality, only use a given concept for the purposes of their own regulation.

This does not mean that LLMs cannot perform functions complementary to CC. LLMs exhibit certain characteristics that make them valuable components of the *cognitive computing* infrastructure. They can synthesize information, detect relationships between concepts, generate summaries, interpretations and hypotheses. In hybrid architecture, they can serve as a component responsible for natural language processing. In this sense, LLMs are not an alternative to CC, but rather a potential complement. They can be used as components of cognitive systems, especially as semantic interfaces and generators of linguistic representations. Their integration with symbolic modules, decision-making agents, procedural memory, and metacognition mechanisms may constitute one of the more interesting directions in the development of artificial intelligence research, especially in applications requiring high cognitive reliability, such as legal analysis and regulatory management.

## 6. CONCLUSIONS

The dynamic growth in the complexity of the legal environment makes classical, manual methods of analysing normative acts insufficient for organisations operating in regulated sectors. The study verified whether combining Large Language Models (LLMs) with the Retrieval-Augmented Generation (RAG) architecture transforms LLMs from tools in their executive version into components of a cognitive infrastructure with enhanced epistemic reliability. It was verified whether the mechanism used in RAG to retrieve relevant document fragments ensures that outputs are constructed basing on actual reference materials, rather than solely on the basis of statistical patterns embedded within the model's parameters. The implementation of such architectures based on locally launched LLMs was proposed, alongside their application as advanced tools for supporting management processes in organisations for which the analysis, interpretation, and implementation of legal requirements are of critical importance. Tools such as `llama.cpp`, Ollama or LM Studio, hosted locally and integrated with an unstructured database, allow for launching large language models even on computers without powerful GPU accelerators, making them especially useful in research, prototyping and installations with limited resources. It was noted that local LLMs, especially when combined with RAG mechanisms and an architecture ensuring auditability and security, constitute a technologically and cognitively mature platform that supports management processes using the analysis of legal acts.

It was demonstrated that application of RAG in the infrastructure of a locally hosted LLM do not replace specialised expert knowledge but create a new infrastructure layer that may strengthen a given organisation's capability to operate in an environment with high regulatory and informational variability. Their implementation enables the transformation of the organisation toward a more efficient, predictive, and automated approach to the interpretation and application of law.

## SUPLEMENTARY MATERIALS

**Table 1.** *Original polish texts generated using production-ready LLMs*

| Bielik | PLLuM |
|---|---|
| **RQ1: Podaj definicję „nieruchomości" w polskim systemie prawnym** | |
| Nieruchomość to grunt, budynek lub część budynku trwale związana z gruntem. W polskim systemie prawnym nieruchomości są podzielone na kilka kategorii: nieruchomości gruntowe (grunty), nieruchomości budynkowe i lokalowe (budynki lub ich części, które stanowią odrębny przedmiot własności). Nieruchomość jest zdefiniowana jako rzecz materialna posiadająca określoną przestrzeń. Właściciel nieruchomości ma pełne prawo do korzystania, rozporządzania i pobierania pożytków ze swojej własności, o ile nie narusza to praw innych osób lub przepisów prawa. Nieruchomości są rejestrowane w księgach wieczystych, które prowadzą informacje o stanie prawnym nieruchomości oraz jej obciążeniach i ograniczeniach. | Nieruchomość to wszystko, co jest trwale związane z gruntem lub stanowi część składową gruntu.<br>Jest to pojęcie używane w prawie cywilnym i odnosi się do wszelkich rzeczy materialnych, które nie są ruchome.<br>Do nieruchomości zaliczamy między innymi budynki, lokale mieszkalne oraz użytkowe, a także części wspólne budynków wielorodzinnych.<br>Nieruchomość może być własnością osób fizycznych lub prawnych i podlegać różnym regulacjom prawnym dotyczącym jej nabywania, zbywania czy dziedziczenia. |
| **RQ2: W jakich ustawach jest to zapisane?** | |
| Definicja nieruchomości w polskim systemie prawnym znajduje się przede wszystkim w Kodeksie cywilnym, który reguluje kwestie związane z własnością i innymi prawami rzeczowymi.<br>Konkretnie, definicja nieruchomości jest zawarta w art. 46 § 1 oraz art. 48 Kodeksu cywilnego:<br>Art. 46 [Pożytki]<br>§ 1. Nieruchomościami są części powierzchni ziemskiej stanowiące odrębny przedmiot własności | Definicja nieruchomości znajduje się w polskim kodeksie cywilnym.<br>Zgodnie z art. 46 § 1 tego aktu prawnego:<br>"Nieruchomościami są części powierzchni ziemskiej stanowiące odrębny przedmiot własności (grunty), jak również budynki trwale z gruntem związane lub części takich budynków, jeżeli na mocy przepisów szczególnych stanowią odrębny od gruntu przedmiot własności." |

| Bielik | PLLuM |
|---|---|
| (grunty), jak również budynki trwale z gruntem związane lub części takich budynków, jeżeli na mocy przepisów szczególnych stanowią odrębny od gruntu przedmiot własności.<br>Art. 48 [Część składowa]<br>Z zastrzeżeniem wyjątków w ustawie przewidzianych, do części składowych gruntu należą w szczególności budynki i inne urządzenia trwale z gruntem związane.<br>Dodatkowo, kwestie dotyczące rejestracji nieruchomości są regulowane przez ustawę o księgach wieczystych i hipotece (z dnia 6 lipca 1982 r.), która określa zasady prowadzenia ksiąg wieczystych oraz ich rolę w obrocie nieruchomościami.<br>Warto także wspomnieć o ustawie o gospodarce nieruchomościami (z dnia 21 sierpnia 1997 r.), która reguluje zasady gospodarowania nieruchomościami Skarbu Państwa oraz jednostek samorządu terytorialnego.<br>Te ustawy razem tworzą ramy prawne dotyczące nieruchomości w Polsce. | Dodatkowo warto wspomnieć, że definicja ta może być rozwijana i uzupełniana przez inne akty prawne oraz orzecznictwo sądowe. |
| **RQ3: Jakie akty prawne rozwinęły tą definicję?** | |
| W polskim systemie prawnym, oprócz Kodeksu cywilnego, ustawy o księgach wieczystych i hipotece oraz ustawie o gospodarce nieruchomościami, istnieje szereg innych aktów prawnych rozwijających definicję nieruchomości w różnych kontekstach:<br>Ustawa prawo budowlane (z dnia 7 lipca 1994 r.) określa zasady projektowania, budowy i utrzymania obiektów budowlanych. Chociaż nie jest to główny akt regulujący kwestię definicji nieruchomości jako takiej, wpływa na charakterystykę i sposób użytkowane nieruchomości przez budynki.<br>Ustawa o planowaniu i zagospodarowaniu przestrzennym (z dnia 27 marca 2003 r.) reguluje procesy planowania przestrzennego. Może mieć wpływ na wartość oraz przeznaczenie nieruchomości w zależności od ich lokalizacji oraz specyfiki urbanistycznej obszarów.<br>Prawo wodne (z dnia 18 lipca 2001 r.) określa status prawny wód i regulacje dotyczące korzystania z gruntów nadbrzeżnych lub położonych blisko rzek czy jezior. Może wpływać na charakterystykę i użytkowanie niektórych rodzajów nieruchomości, szczególnie tych związanych z wodą.<br>Ustawa o gospodarce komunalnej (z dnia 8 marca 1990 r.) dotyczy działalności gmin w zakresie gospodarki nieruchomościami komunalnymi. To może mieć znaczenie przy definiowaniu lub wykorzystywaniu pewnych typów nieruchomości samorządowych.<br>Te akty prawne nie modyfikują podstawowej definicji nieruchomości zawartej w Kodeksie cywilnym, ale rozwijają ją w kontekście specyficznych sytuacji oraz rodzajów nieruchomości. | Definicja nieruchomości zawarta w kodeksie cywilnym jest podstawowym źródłem prawa w tej kwestii.<br>Jednakże istnieją również inne akty prawne, które rozwijają tę definicję i wprowadzają dodatkowe regulacje dotyczące nieruchomości. Przykładem takiego aktu może być ustawa o gospodarce nieruchomościami z dnia 21 sierpnia 1997 roku, która określa zasady obrotu nieruchomościami w Polsce oraz procedury związane z ich podziałem, łączeniem czy wywłaszczeniem.<br>Ponadto warto wspomnieć o ustawie Prawo budowlane, która reguluje kwestie związane ze wznoszeniem i utrzymaniem obiektów budowlanych oraz wpływa na sposób postrzegania nieruchomości jako całości funkcjonalnej. |
| **RQ4: Nigdzie więcej?** | |

| Bielik | PLLuM |
|---|---|
| W polskim systemie prawnym główne regulacje dotyczące nieruchomości znajdują się przede wszystkim w Kodeksie cywilnym, ustawie o księgach wieczystych i hipotece oraz ustawie o gospodarce nieruchomościami. Te akty prawne stanowią podstawę dla definicji, rejestracji i obrotu nieruchomościami.<br>Jeśli chodzi o szczegółowe kwestie dotyczące np. zarządzania nieruchomościami wspólnymi w budynkach mieszkalnych, mogą one być regulowane przez ustawę o własności lokali (z dnia 24 czerwca 1994 r.). Jednakże definicja samej nieruchomości jako taka jest przede wszystkim zawarta we wspomnianych wyżej aktach prawnych.<br>Warto zaznaczyć, że mogą istnieć także inne ustawy szczególne (lex specialis), które regulują specyficzne aspekty dotyczące nieruchomości w określonych kontekstach. Jednak podstawowa definicja i ogólne zasady są zawarte głównie we wspomnianych wyżej aktach prawnych. | Definicja nieruchomości może być również rozwijana przez orzecznictwo sądowe.<br>W praktyce oznacza to, że w przypadku sporów dotyczących statusu prawnego danej rzeczy lub jej części decydujące znaczenie mają wyroki wydawane przez polskie sądy.<br>Ponadto warto wspomnieć o poglądach doktryny prawa cywilnego, które mogą wpływać na interpretację pojęcia nieruchomości i jego praktyczne zastosowanie w konkretnych sytuacjach życiowych czy biznesowych. |

**Table 2.** *Original polish texts generated using LLMs extended with RAG*

| Bielik+RAG | PLLuM+RAG |
|---|---|
| **RQ1: Podaj definicję „nieruchomości" w polskim systemie prawnym** | |
| W polskim systemie prawnym nieruchomość definiowana jest jako część powierzchni ziemskiej, stanowiąca odrębny przedmiot własności. Jest to podstawowy element regulacji dotyczących prawa rzeczowego i ma kluczowe znaczenie dla obrotu gospodarczego oraz systemu finansowego kraju. Nieruchomość może być gruntowa, budynkowa lub lokalowa (mieszkalna bądź użytkowa). Grunty to tereny niezabudowane lub zabudowane obiektami trwale związanymi z podłożem. Budynki mogą stanowić samodzielną nieruchomość, jeśli są wzniesione na gruncie będącym przedmiotem użytkowania wieczystego lub gdy prawo do budynku wynika z odrębnej umowy. Natomiast lokale (mieszkalne i użytkowe) to wydzielone części budynków, które mogą być przedmiotem obrotu prawnego niezależnie od gruntu czy pozostałej części budynku. Definicja ta ma fundamentalne znaczenie dla wszelkich transakcji związanych z nieruchomościami oraz jest podstawą do ustanawiania i ochrony praw własnościowych. | W polskim systemie prawnym, pojęcie nieruchomości jest zdefiniowane w Kodeksie cywilnym (K.c.). Zgodnie z art. 46 § 1 K.c., "nieruchomościami są części powierzchni ziemskiej stanowiące odrębny przedmiot własności (grunty), jak również budynki trwale z gruntem związane lub części takich budynków, jeżeli na mocy przepisów szczególnych stanowią odrębny od gruntu przedmiot własności." Definicja ta ma charakter ogólny i stanowi podstawę dla bardziej szczegółowych regulacji dotyczących nieruchomości. Warto zauważyć, że w prawie polskim wyróżnia się różne rodzaje nieruchomości: Grunty - wszelkie części powierzchni ziemskiej stanowiące odrębny przedmiot własności (np. działki budowlane, rolne). Budynki trwale z gruntem związane - konstrukcje wzniesione na gruncie, które nie mogą być łatwo przenoszone. Części takich budynków - w przypadku gdy przepisy szczególne tak stanowią (np. lokale mieszkalne wyodrębnione jako odrębna własność). Nieruchomość może być także ruchoma, jeśli stanowi część składową nieruchomości gruntowej lub budynkowej. Ważnym elementem definicji jest trwałe związanie z gruntem - oznacza to, że nie można łatwo przenieść danego obiektu w inne miejsce bez zniszczenia jego wartości użytkowej. |
| **RQ2: W jakich ustawach jest to zapisane?** | |
| Definicja "nieruchomości" oraz regulacje związane z nieruchomościami są zawarte głównie w dwóch podstawowych aktach prawnych: Kodeksie cywilnym (KC) i Ustawie o gospodarce nieruchomościami.<br>- Kodeks Cywilny – Art. 46 § 1 KC definiuje nieruchomość jako część powierzchni ziemskiej, | Odpowiedź na pytanie o przepisy dotyczące nieruchomości w polskim systemie prawnym wymaga szczegółowego odniesienia się do kilku kluczowych aktów legislacyjnych. Poniżej przedstawiono najważniejsze z nich, wraz z ich odpowiednimi artykułami i paragrafami: Kodeks cywilny (K.c.) Artykuł 46 § 1: "Nieruchomościami |

| Bielik+RAG | PLLuM+RAG |
| --- | --- |
| stanowiącą odrębny przedmiot własności. Dodatkowo, art. 235 KC określa budynki i inne urządzenia trwale związane z gruntem jako część składową nieruchomości (chyba że przepisy szczególne stanowią inaczej).<br>- Ustawa o gospodarce nieruchomościami – Ta ustawa reguluje kwestie związane z wywłaszczeniem, podziałem i scalaniem nieruchomości oraz ich sprzedażą przez Skarb Państwa lub jednostki samorządu terytorialnego. Choć nie definiuje bezpośrednio pojęcia "nieruchomość", odnosi się do różnych aspektów zarządzania i obrotu nieruchomościami, co jest istotne w kontekście prawa cywilnego.<br>Dodatkowo, szczegółowe regulacje dotyczące poszczególnych typów nieruchomości (np. gruntowych, budynkowych, lokalowych) mogą znajdować się w innych ustawach i rozporządzeniach wykonawczych, takich jak Prawo geodezyjne i kartograficzne czy Ustawa o własności lokali. | są części powierzchni ziemskiej stanowiące odrębny przedmiot własności (grunty), jak również budynki trwale z gruntem związane lub części takich budynków, jeżeli na mocy przepisów szczególnych stanowią odrębny od gruntu przedmiot własności." Prawo geodezyjne i kartograficzne Artykuł 2 pkt. 8: Definicja działki ewidencyjnej jako jednostki powierzchniowej wykorzystywanej do opisu stanu prawnego nieruchomości. Ustawa o gospodarce nieruchomościami Rozdział I, art. 4: Zawiera ogólne zasady dotyczące obrotu i zarządzania nieruchomościami w Polsce, a także definicję nieruchomości jako przedmiotu prawa własności. Ustawa o księgach wieczystych i hipotece Artykuł 2 pkt. 1: Definiuje nieruchomość jako grunt wraz z częściami składowymi oraz budynkami trwale związanymi z tym gruntem, jeśli stanowią odrębny przedmiot własności. Ustawa o planowaniu i zagospodarowaniu przestrzennym Artykuł 2 pkt. 13: Określa definicję terenu jako nieruchomości przeznaczonej pod zabudowę lub użytkowanej w określony sposób, zgodnie z miejscowym planem zagospodarowania przestrzennego. Ustawa o ochronie gruntów rolnych i leśnych Artykuł 4 pkt. 1: Wprowadza pojęcie nieruchomości jako gruntu przeznaczonego do rolnictwa lub leśnictwa, podlegającego szczególnej ochronie prawnej. Ustawa o podatkach i opłatach lokalnych Artykuł 2 ust. 1: Definiuje przedmiot opodatkowania w postaci nieruchomości jako gruntów, budynków lub ich części, a także budowli związanych z prowadzeniem działalności gospodarczej. Każda z tych ustaw zawiera przepisy regulujące różne aspekty dotyczące nieruchomości - od definicji, przez zasady obrotu i użytkowania, po kwestie podatkowe oraz ochronę gruntów rolnych i leśnych. |
| **RQ3: Jakie akty prawne rozwinęły tą definicję?** | |
| Definicja nieruchomości zawarta w Kodeksie cywilnym (KC) została rozwinięta i uszczegółowiona przez szereg innych aktów prawnych, które regulują różne aspekty związane z nieruchomościami. Oto najważniejsze z nich:<br>- Ustawa o księgach wieczystych i hipotece – Reguluje prowadzenie ksiąg wieczystych, które są podstawowym rejestrem praw do nieruchomości. Ustawa ta określa zasady wpisów i wykreśleń oraz ochronę praw rzeczowych na nieruchomościach.<br>- Ustawa o gospodarce nieruchomościami – Określa zasady gospodarowania nieruchomościami publicznymi (Skarbu Państwa i jednostek samorządu terytorialnego), w tym wywłaszczenia, podziału, scalania oraz sprzedaży.<br>- Prawo geodezyjne i kartograficzne – Reguluje kwestie związane z ewidencją gruntów i budynków (kataster nieruchomości), co ma kluczowe znaczenie dla ustalania stanu prawnego nieruchomości.<br>- Ustawa o własności lokali – Określa zasady ustanawiania i zarządzania odrębną własnością lokali (mieszkalnych i użytkowych), co jest istotne dla | Rozwinięcie definicji nieruchomości w polskim systemie prawnym wynika z szeregu aktów legislacyjnych, które dostosowują ogólne zasady zawarte w Kodeksie cywilnym do specyfiki różnych typów nieruchomości oraz regulują szczegółowe aspekty ich użytkowania i zarządzania. Poniżej przedstawiono kluczowe akty prawne, które rozszerzyły definicję nieruchomości: Ustawa o gospodarce nieruchomościami (UGN) Rozdział I: Ustanawia podstawy gospodarki nieruchomościami w Polsce. Artykuł 4 pkt. 1: Zawiera ogólną definicję nieruchomości gruntowej jako części powierzchni ziemskiej stanowiącej odrębny przedmiot własności. Ustawa o księgach wieczystych i hipotece (UKWH) Artykuł 2 pkt. 1: Określa, że nieruchomość może składać się z gruntu wraz z częściami składowymi oraz budynkami trwale związanymi z tym gruntem. Ustawa o planowaniu i zagospodarowaniu przestrzennym Artykuł 2 pkt. 13: Wprowadza pojęcie terenu jako nieruchomości przeznaczonej pod zabudowę lub użytkowanej w określony sposób, zgodnie z miejscowym planem zagospodarowania |

| Bielik+RAG | PLLuM+RAG |
|---|---|
| nieruchomości lokalowych.<br>- Ustawa o planowaniu i zagospodarowaniu przestrzennym – Reguluje kwestie związane z przeznaczeniem terenów, co wpływa na możliwość korzystania z nieruchomości (np. budowa, użytkowanie).<br>- Ustawa o ochronie gruntów rolnych i leśnych – Określa zasady ochrony gruntów rolnych i leśnych oraz ich przekształcania na cele nierolnicze lub nieleśne.<br>Te akty prawne, w połączeniu z Kodeksem cywilnym, tworzą kompleksowy system regulacji dotyczących nieruchomości w Polsce. | przestrzennego. Ustawa Prawo budowlane Artykuł 3 pkt. 2: Definiuje budynki oraz ich części jako odrębne obiekty budowlane stanowiące nieruchomość lub jej część składową. Ustawa o podatkach i opłatach lokalnych Rozdział III, art. 1a ust. 1: Wskazuje na różnorodność przedmiotów opodatkowania w postaci gruntów, budynków oraz ich części, a także budowli związanych z prowadzeniem działalności gospodarczej. Ustawa o rolnictwie Artykuł 4 pkt. 1: Określa nieruchomość rolną jako grunt przeznaczony do prowadzenia działalności rolnej, podlegający szczególnym regulacjom dotyczącym obrotu i użytkowania. Ustawa o ochronie środowiska Artykuł 3 pkt. 48: Zawiera definicję nieruchomości jako obszaru, który może być przedmiotem działań mających wpływ na środowisko naturalne. Każda z tych ustaw wnosi istotny wkład w rozwinięcie ogólnej definicji nieruchomości zawartej w Kodeksie cywilnym, dostosowując ją do specyfiki różnych obszarów prawa oraz potrzeb praktycznych związanych z zarządzaniem i obrotem nieruchomościami. |
| **RQ4: Nigdzie więcej?** | |
| W polskim systemie prawnym podstawowe definicje i regulacje dotyczące nieruchomości są zawarte głównie w Kodeksie cywilnym oraz kilku kluczowych ustawach, które szczegółowo rozwijają tę tematykę. Wymienione przeze mnie akty prawne (KC, ustawa o księgach wieczystych i hipotece, ustawa o gospodarce nieruchomościami, prawo geodezyjne i kartograficzne, ustawa o własności lokali, ustawa o planowaniu i zagospodarowaniu przestrzennym oraz ustawa o ochronie gruntów rolnych i leśnych) stanowią fundament prawny dla kwestii związanych z nieruchomościami.<br>Oprócz tego, istnieją również inne przepisy branżowe lub szczegółowe rozporządzenia wykonawcze, które mogą odnosić się do konkretnych aspektów zarządzania nieruchomościami (np. podatki od nieruchomości, prawo budowlane w zakresie dotyczącym budynków). Jednak główne ramy prawne dla definicji i regulacji nieruchomości są już kompleksowo ujęte we wspomnianych aktach.<br>Jeśli chodzi o bardziej szczegółowe lub specjalistyczne kwestie, mogą one być rozwijane w doktrynie prawniczej (orzecznictwo sądów, komentarze), ale same akty normatywne nie zawierają już dodatkowych definicji poza tymi podstawowymi. | W polskim systemie prawnym definicja nieruchomości jest rozwijana nie tylko w aktach legislacyjnych, ale również znajduje swoje odzwierciedlenie w licznych orzeczeniach sądowych oraz interpretacjach organów administracyjnych. Poniżej przedstawiono kluczowe źródła, które dodatkowo rozszerzają tę definicję: Orzecznictwo Sądów Najwyższych Wiele spraw dotyczących nieruchomości rozpatrywanych przez Sąd Najwyższy (SN) przyczyniło się do doprecyzowania pojęcia nieruchomości w polskim prawie. Przykłady istotnych orzeczeń SN to m.in.: Uchwała z dnia 15 lutego 2019 roku (sygnatura III CZP 83/18): Wyjaśnia, że budynek trwale związany z gruntem może stanowić odrębną nieruchomość tylko wtedy, gdy przepisy szczególne tak stanowią. Wyrok z dnia 4 lutego 2022 roku (sygnatura II CSKP 103/22): Potwierdza, że części budynku nie mogą być uznane za nieruchomości bez wyraźnego uregulowania ustawowego. Interpretacje organów administracyjnych Ministerstwo Rozwoju i Technologii oraz Ministerstwo Finansów regularnie wydają interpretacje dotyczące definicji nieruchomości w kontekście różnych aspektów prawa, takich jak podatki czy planowanie przestrzenne. Przykładowe stanowiska obejmują: Interpretacja indywidualna MF z dnia 15 stycznia 2023 roku (nr 0114-KDIP1-2.4012.719.2022.2.MC): Wyjaśnia, że dla celów podatkowych nieruchomość obejmuje nie tylko grunt i budynek, ale również związane z nimi części składowe. Stanowisko MRiT z dnia 8 lutego 2024 roku: Podkreśla znaczenie miejscowych planów zagospodarowania przestrzennego w definiowaniu przeznaczenia terenu jako nieruchomości budowlanej lub innej. Warto również zwrócić uwagę na prace doktrynalne oraz |

| Bielik+RAG | PLLuM+RAG |
| --- | --- |
| | publikacje naukowe, które analizują pojęcie nieruchomości z perspektywy prawa cywilnego, administracyjnego i podatkowego. Dzięki szerokiemu spektrum źródeł prawnych definicja nieruchomości w polskim systemie prawnym pozostaje elastyczna i dostosowana do zmieniających się realiów gospodarczych oraz społecznych. |